\documentclass[10pt,conference]{IEEEtran}
\IEEEoverridecommandlockouts

\usepackage{cite}
\usepackage{amsmath,amssymb,amsfonts}
\usepackage{algorithmic}
\usepackage{graphicx}
\usepackage{textcomp}
\usepackage{xcolor}

\def\BibTeX{{\rm B\kern-.05em{\sc i\kern-.025em b}\kern-.08em
    T\kern-.1667em\lower.7ex\hbox{E}\kern-.125emX}}
\begin{document}

\title{
    Mixed Precision PointPillars for Efficient 3D Object Detection with TensorRT
}

\author{
    \IEEEauthorblockN{1\textsuperscript{st} Ninnart Fuengfusin}
    \IEEEauthorblockA{
        \textit{Advanced Mobility Research Institute} \\
        \textit{Kanazawa University}\\
        Ichikawa, Japan \\
        ninnart@se.kanazawa-u.ac.jp
    }
    \and

    \IEEEauthorblockN{2\textsuperscript{nd} Keisuke Yoneda}
    \IEEEauthorblockA{
        \textit{Advanced Mobility Research Institute} \\
        \textit{Kanazawa University}\\
        Ichikawa, Japan \\
        k.yoneda@staff.kanazawa-u.ac.jp
    }
    \and

    \IEEEauthorblockN{3\textsuperscript{rd} Naoki Suganuma}
    \IEEEauthorblockA{
        \textit{Advanced Mobility Research Institute} \\
        \textit{Kanazawa University}\\
        Ichikawa, Japan \\
        suganuma@se.kanazawa-u.ac.jp
    }
}
\maketitle

\begin{abstract}
LIDAR 3D object detection is one of the important tasks for autonomous vehicles.
Ensuring that this task operates in real-time is crucial. Toward this, model
quantization can be used to accelerate the runtime. However, directly applying model
quantization often leads to performance degradation due to LIDAR's wide numerical
distributions and extreme outliers. To address the wide numerical distribution, we
proposed a mixed precision framework designed for PointPillars. Our framework first
searches for sensitive layers with post-training quantization (PTQ) by quantizing one
layer at a time to 8-bit integer (INT8) and evaluating each model for average precision
(AP). The top-k most sensitive layers are assigned as floating point (FP). Combinations
of these layers are greedily searched to produce candidate mixed precision models, which
are finalized with either PTQ or quantization-aware training (QAT). Furthermore, to
handle outliers, we observe that using a very small number of calibration data reduces
the likelihood of encountering outliers, thereby improving PTQ performance. Our methods
provides mixed precision models without training in the PTQ pipeline, while our QAT
pipeline achieves the performance competitive to FP models. With TensorRT deployment,
our mixed precision models offer less latency by up to 2.538 times compared to FP32
models.
\end{abstract}


\section{Introduction}
\label{sec:intro}
LIDAR 3D object detection has enabled autonomous cars to recognize the surrounding
environment. To ensure that the task performs accurately and operates in real-time is
crucial to success. However, autonomous cars are often equipped with edge devices with
limited computing capacity and memory. Hence, the deployment of 3D object detection
models has become the challenge.

Toward this, PointPillars \cite{lang2019pointpillars} are promising candidate models due
to their low runtime latency compared to other models, such as point-based models
\cite{qi2017pointnet} and voxel-based models \cite{zhou2018voxelnet}. This is because
PointPillars process 3D point cloud into 2D pseudo-images, which can be fed directly
into 2D convolutions.  This processing helps avoid the use of expensive 3D convolutions
\cite{zhou2018voxelnet} and complex point sampling \cite{qi2017pointnet++}.

Furthermore, model quantization promises to reduce latency by constraining the expensive
32-bit floating point (FP32) to a lower bit-width datatype, such as 8-bit integer
(INT8). This enables low-precision arithmetic that is faster and requires lower memory
bandwidths.

To quantize a model, post-training quantization (PTQ) is attractive, compared to
quantization aware training (QAT). Since PTQ does not require any training, it
requires only a small set of data for calibration by performing forward propagations and
tracking statistical information. However, directly applying PTQ INT8 quantization to
PointPillars causes a dramatic performance drop \cite{zhou2024lidar}. QAT may reduce
the performance drop from INT8 quantization; however, there is still a performance gap
compared to the FP32 model, as shown in Table \ref{tab:ptq-qat}.

\begin{table}[htbp]
    \caption{
        INT8 PTQ and QAT with PointPillars. The models were evaluated on the KITTI 3D
        object detection dataset \cite{geiger2012we}. The 3D average precision over 11
        and 40 recall values (AP11, AP40) are reported over three difficulties.
    }
    \begin{center}
    \begin{tabular}{|c|c|c|c|c|c|c|}
        \hline
        \textbf{Data Type} &
        \multicolumn{3}{c|}{\textbf{Car AP11}} &
        \multicolumn{3}{c|}{\textbf{Car AP40}} \\
        \cline{2-7}
        & \textbf{Easy} & \textbf{Mod.} & \textbf{Hard}
        & \textbf{Easy} & \textbf{Mod.} & \textbf{Hard}\\
        \hline
        FP32 & \textbf{86.35} & \textbf{76.95} & \textbf{75.46} & \textbf{87.13} & \textbf{78.49} & \textbf{75.84}\\
        \hline
        PTQ INT8 & 27.29 & 23.63 & 20.18 & 25.35 & 21.43 & 18.66 \\
        \hline
        QAT INT8 & 75.06 & 64.64 & 61.48 & 77.93 & 64.7 & 60.56 \\
        \hline
    \end{tabular}
    \label{tab:ptq-qat}
    \end{center}
\end{table}

In this work, instead of introducing novel architectures \cite{zhou2025pillarhist} or
quantization schemes \cite{zhou2024lidar}, we propose using a mixed precision based
method by quantizing less sensitive layers to INT8 and more sensitive layers to
16-bit floating points (FP16).

To find sensitive layers, PTQ is used to quantize one of layers to INT8 and evaluate
the 3D average precision over 40 recall values (AP40). With PTQ, the sensitive search is
performed without additional training. Layers with low AP40 scores are denoted as the
high-sensitivity layers.

\begin{figure}[htbp]
    \centerline{
        \includegraphics[scale=0.4]{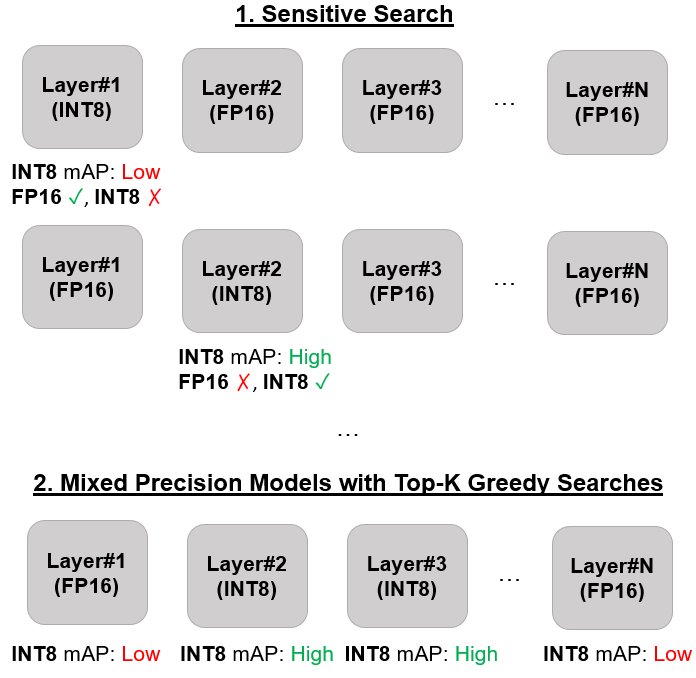}
    }
    \caption{
        Overview of our proposed method. First, sensitive layers are identified by
        quantizing one layer at a time to INT8 using PTQ. Second, a greedy search is
        used to discover mixed precision models where these sensitive layers are
        replaced with FP16 layers.
    }
    \label{fig:overview}
\end{figure}

After sensitive layers are identified, a greedy search is used to discover combinations
of INT8 sensitive layers to assign with the FP16 datatype. Finally, these layer
combinations are calibrated again with either PTQ or QAT. With mixed precision and
greedy search, we demonstrate that conventional quantization algorithms can operate
without any modifications. The overview of these methods is visualized in Fig.
\ref{fig:overview}

Furthermore, for PTQ, we observe that using a very small number of calibration data
leads to better performance of models. This is because the fewer calibration samples
used, the lower the chance of encountering extreme outliers. Therefore, this reduces
rounding errors due to wider quantization step sizes.

To ensure that our mixed precision models can be deployed, our method complies with
TensorRT \cite{tensorrt} requirements. We provide extensive evaluations of TensorRT's
runtime latency across two devices: Jetson Orin and RTX 4070Ti.

In this work, we focus on using only FP16 and INT8 datatypes due to their wide support
across both old and modern hardware. With these approaches, our method aims to minimize
AP40 loss from INT8 quantization while providing the lower runtime latency compared to
other models.

To the best of our knowledge, our main contributions in this paper are listed as
follows:

\begin{itemize}
    \item We propose a mixed precision framework designed for PointPillars and TensorRT.
    Instead of using only latency as the metric, as commonly used with TensorRT, our
    method instead uses model performance to search for optimal layer combinations
    instead.
    \item Our framework uses PTQ to identify top-$k$ sensitive layers and utilizes
    greedy searches to find top-$k$ optimal combinations of mixed precision layers for
    further using with either PTQ or QAT.
    \item We observe a negative correlation between PTQ model performance and the number
    of calibration data. To enhance the performance of PTQ models, we propose using a
    very small number of calibration data to avoid extreme outliers.
\end{itemize}

\section{Related Works}
\label{sec:related}
In this section, we describe the related works that lie at the intersection between 3D
object detection models and quantization.

Quantizing 3D object detection models is challenging due to high sparsity, extreme
outliers, and wide numerical distributions \cite{zhou2024lidar, zhou2025pillarhist}. To
perform PTQ on 3D object detection model, LIDAR-PTQ \cite{zhou2024lidar} was proposed
with a new data calibration technique that accounts for the sparsity and introduces a
loss function that minimizes the prediction differences between FP32 and INT8 models.
With the extra-loss term, LIDAR-PTQ is in between PTQ and QAT, as it requires training
to optimize quantization parameters such as scaling factors and zero points, while
requiring only unlabeled data.

On the other hand, PillarHist \cite{zhou2025pillarhist} was proposed with a novel
pillar encoder that stabilizes the numerical distribution of variables within the pillar
encoder, thereby enhancing PointPillars robustness to INT8 quantization.

Stanisz \textit{et al.} \cite{stanisz2020optimisation} demonstrated the use of QAT with
a PyTorch quantization framework, \texttt{brevitas} \cite{brevitas} to quantize
PointPillars with different fixed point data precision across sub-modules. With
empirical experiments, Stanisz \textit{et al.} identified a combination between 2-, 4-,
8-bit integer sub-modules that achieves a 16x lower memory consumption, while average
precision is reduced by at most 9\% across all classes.

Compared to previous works, our work addresses the same research problems as LIDAR-PTQ
\cite{zhou2024lidar} and PillarHist \cite{zhou2025pillarhist}. However, our approach
differs by using mixed precision models to handle numerical distribution stabilization
and using a very small number of calibration data to handle extreme outliers.

For Stanisz \textit{et al.} \cite{stanisz2020optimisation} work, our works differs in
that, instead of manually searching for data precision across sub-modules, our works
uses PTQ to find sensitive layers and assign the data precision with. For
deployment, \texttt{brevitas} \cite{brevitas}, which uses asymmetric quantization by
default, may cause an incompatibility to TensorRT deployment, as TensorRT only
supports symmetric quantization \cite{tensorrt}. With TensorRT, the precision of ReLU
outputs is reduced by half and this can cause greater losses compared to Stanisz
\textit{et al.}.

In terms of mixed precision, TensorRT supports a mixed precision scheme; however, it
finds a layer combination that minimizes latency, whereas our proposed method focuses on
maximizing model accuracy. Furthermore, we observed that to enable this scheme, all
layers should first be quantized to INT8, allowing TensorRT to choose between INT8 and
other datatypes. However, relying on this selection can lead to degradation in model
accuracy due to accumulated quantization losses from all layers.

\section{Proposed Method}
\label{sec:proposed}
In this section, we first describe the prerequisites of PointPillars, INT8 quantization
and then discuss our proposed methods, mixed precision PointPillars.

\subsection{Prerequisites}
\subsubsection{PointPillars}
PointPillars \cite{lang2019pointpillars} proposes a novel pillar encoder that processes
3D point clouds into vertical columns called pillars. These pillars are further
processed with PointNet \cite{qi2017pointnet} and scattered into 2D pseudo-images. The
2D pseudo-images are fed into 2D convolutional neural network (CNN) backbone, which
produces features for bounding box prediction with the detection head.

Based on these processes, \texttt{MMDetection3D} \cite{mmdet3d2020} further decomposes
PointPillars into five sub-modules: voxel encoder, middle encoder, backbone, neck, and
bbox head. An overview of this architecture is shown in Fig. \ref{fig:overview}. The
voxel encoder processes the 3D point cloud into pillars and further encodes them using
PointNet. The middle encoder scatters the voxel encoder output into 2D pseudo-images.
The backbone processes these pseudo-images with 2D CNN, and the neck aggregates
multi-scale features, which are then consumed by the bbox head to produce the 3D
bounding boxes prediction.

\subsubsection{INT8 Quantization and TensorRT}
INT8 quantization is formulated as the affine transformation in \eqref{eq:quantization},
which maps a floating point value, $x$ to its corresponding integer value, $x_{q}$.
Here, $\lfloor\cdot\rceil$ denotes rounding half to even and the
$\operatorname{clamp}(x, q_{min}, q_{max})$ function clips $x$ values to the range of
$[q_{min}, q_{max}]$.

\begin{equation}
    x_{q} = \operatorname{clamp}\bigl(\lfloor \frac{x}{s} \rceil + z,\; q_{\min},\; q_{\max}\bigr)
    \label{eq:quantization}
\end{equation}

To comply with TensorRT requirements \cite{tensorrt} that supports only symmetric
quantization, we set $z = 0$, $q_{min} = -128$, and $q_{max} = 127$ for INT8
quantization. Therefore, \eqref{eq:quantization} becomes \eqref{eq:quantization2}.

\begin{equation}
    x_{q} = \operatorname{clamp}\bigl(\lfloor \frac{x}{s} \rceil,\; q_{\min},\; q_{\max}\bigr)
    \label{eq:quantization2}
\end{equation}

For PTQ, min-max calibration provides better performance compared to entropy calibration
for the LIDAR 3D object detection task \cite{zhou2024lidar}. Therefore, the min-max
calibration is chosen and used to solve for $s$ by \eqref{eq:quantization3}, where
$x_{min}$ and $x_{max}$ are the minimum and maximum values observed during data
calibration.

\begin{equation}
    s = \frac{\operatorname{max}(\operatorname{abs}(x_{min}), \operatorname{abs}(x_{max}))}{127}
    \label{eq:quantization3}
\end{equation}

\subsection{Mixed Precision PointPillars}
In this section, we describe our proposed method, mixed precision PointPillars. We first
describe how to discover sensitive layers and how to find suitable layer combinations
for mixed precision models to use with final PTQ and QAT. After that, we discuss the
concept behind our proposed method design, such as PointPillars latency for each layer
across datatype and the correlation between the number of calibration data and model
performance.

\subsubsection{PTQ Sensitive Layer Search}
\label{sec:sensitive}
To identify sensitive layers, PTQ is performed by quantizing one layer at a time to
INT8 and evaluating for AP40. Then, the top-$k$ sensitive layers are selected to
generate layer combinations for the candidate model.

To formalize, let $D_{train}$, $D_{val}$, and $D_{cal}$ denote the training, validation,
and calibration datasets, respectively, where $D_{cal} \subseteq D_{train}$. Let the
3D average precision metric be defined as the function of $AP40: (\theta, D_{val})
\rightarrow \mathbb{R}$.

To measure sensitivity, we construct the partially quantized model $\theta^{l}$ with a
single $l$-th INT8 layer created by PTQ, using $D_{cal}$. Our objective is to search for
top-$k$ layers that yield the lowest validation performance. This statement is
formulated as \eqref{eq:sensitive}, where  $\mathcal{L} = \{1, \dots, L\}$ is the set of
all layer indices and $S$ is the set of top-$k$ sensitive layer indices.

\begin{equation}
    S^{*}
    = \underset{S \subseteq \mathcal{L},\, |S| = k}{\arg\min}
      \sum_{l \in S} AP40\bigl(\theta^{l}, D_{\text{val}}\bigr).
    \label{eq:sensitive}
\end{equation}

After top-$k$ sensitive layers $S^{*}$ are found by searching across layers, with PTQ,
this process does not require a significant amount of time. However, to find the most
optimal combination of layers requires $\binom{L}{k} = \frac{L!}{k!(L - k)!}$
combinations to search. With large $L$, this becomes combinatorially expensive.

To address this issue, we adopt a greedy search by ranking the most sensitive layers
from the most to the least sensitive. Then, we construct candidate models by selecting
the top-$1$ to top-$k$ sensitive layers to assign as the FP16 layers of the INT8 models.
With this approach, only $k$ combinations are required for the search.

After the search, once the candidate mixed precision models are found, we perform the
final PTQ or QAT to calibrate or train the mixed precision models. With the sensitive layers
replaced, our models can perform without issues caused by sensitive layers.

\subsubsection{INT8 and Latency}
\label{sec:int8-latency}
To minimize latency, our proposed method is designed based on the promise that INT8 models
deliver lower runtime compared to other datatypes. Therefore, by ensuring that almost all
layers are INT8 and only a few sensitive layers are FP16, the overall latency of
PointPillars model can be minimized.

To validate this statement, we convert PointPillars models to FP32, FP16, and INT8 with
\texttt{trtexec}, a command line tool to TensorRT \cite{tensorrt}. With this,
we benchmarked the latencies across layers, as shown in Table \ref{tab:latency}.

\begin{table}[htbp]
    \caption{
        Latency for each weight layer within PointPillars across three datatypes on the
        to Jetson Orin. Each layer is fused with its corresponding batch normalization
        and ReLU.
    }
    \begin{center}
    \begin{tabular}{|c|c|c|c|c|}
        \hline
        \textbf{Index} & \textbf{Layer Name} & \multicolumn{3}{c|}{\textbf{Latency (ms)}} \\
        \cline{3-5}
        & & \textbf{FP32} & \textbf{FP16} & \textbf{INT8} \\
        \hline
        1 & voxel\_encoder.pfn\_layers.0.linear & 0.8624 & \textbf{0.8542} & 0.8655  \\
        \hline
        2 & backbone.blocks.0.0                 & 1.382 & 0.5614 & \textbf{0.3759} \\
        \hline
        3 & backbone.blocks.0.3                 & 1.369 & 0.5586 & \textbf{0.3667} \\
        \hline
        4 & backbone.blocks.0.6                 & 1.369	& 0.5551 & \textbf{0.3663} \\
        \hline
        5 & backbone.blocks.0.9                 & 1.369	& 0.5548 & \textbf{0.3661} \\
        \hline
        6 & backbone.blocks.1.0                 & 0.6487 & 0.3045 & \textbf{0.2454} \\
        \hline
        7 & backbone.blocks.1.3                & 1.014	& 0.4941 & \textbf{0.1749} \\
        \hline
        8 & backbone.blocks.1.6                & 1.011	& 0.4926 & \textbf{0.2771} \\
        \hline
        9 & backbone.blocks.1.9                 & 1.01	& 0.4925 & \textbf{0.276} \\
        \hline
        10 & backbone.blocks.1.12                & 1.011 & 0.4915 & \textbf{0.2762} \\
        \hline
        11 & backbone.blocks.1.15                & 1.011 & 0.4926 & \textbf{0.2759} \\
        \hline
        12 & backbone.blocks.2.0                & 0.5441 & \textbf{0.2559} & 0.2764 \\
        \hline
        13 & backbone.blocks.2.3                & 0.9886 & 0.4695 & \textbf{0.4104} \\
        \hline
        14 & backbone.blocks.2.6               & 0.9879 & 0.4688 & \textbf{0.1483} \\
        \hline
        15 & backbone.blocks.2.9                & 0.9894 & 0.4687 & \textbf{0.2554} \\
        \hline
        16 & backbone.blocks.2.12                & 0.9874 & 0.4696 & \textbf{0.2552} \\
        \hline
        17 & backbone.blocks.2.15                & 0.9882 & 0.4694 & \textbf{0.2553} \\
        \hline
        18 & neck.deblocks.0.0                  & 0.6957 & 0.3634 & \textbf{0.2553} \\
        \hline
        19 & neck.deblocks.1.0                  & 1.051	& 0.5117 & \textbf{0.2553} \\
        \hline
        20 & neck.deblocks.2.0                  & 1.666	& 0.7109 & \textbf{0.5723}   \\
        \hline
        21 & bbox\_head.conv\_dir\_cls               & 0.9324 & 0.3979 & \textbf{0.2423} \\
        \hline
        22 & bbox\_head.conv\_reg          & 0.9781 & \textbf{0.4527} & 0.4807 \\
        \hline
        23 & bbox\_head.conv\_cls               & 0.7473 & 0.4082 & \textbf{0.2545}  \\
        \hline
    \end{tabular}
    \label{tab:latency}
    \end{center}
\end{table}

From the table, in general, almost all INT8 layers promise a faster runtime compared
FP32 and FP16 layers. Therefore, in terms of individual layer, INT8 layers provide lower
runtime than FP32 and FP16 layers.

\subsubsection{Data Calibration and Performance}
LIDAR data is highly sparse and exhibits wide numerical distributions
\cite{zhou2024lidar}. These properties often result in extreme outliers or values with
large magnitude. In min-max calibration, these outliers determine $x_{max}$, causing the
$s$ scaling factor or step size in \eqref{eq:quantization3} to increase. Consequently,
the rounding error increases while the clipping error decreases.

\begin{figure}[htbp]
    \centerline{
        \includegraphics[scale=0.5]{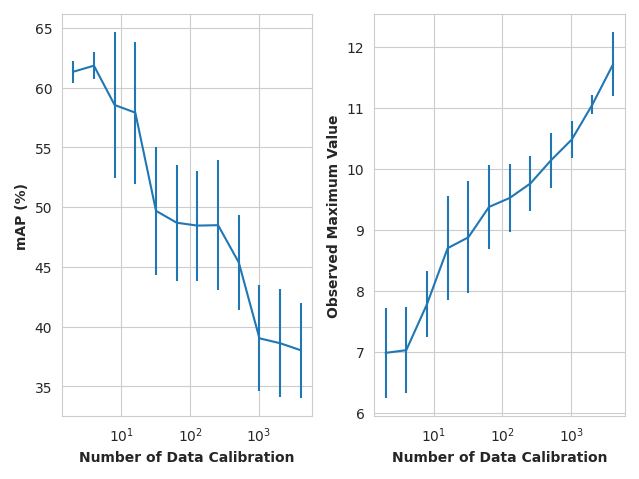}
    }
    \caption{
        Error bars indicate mean and standard deviation across five random seeds.
        \textbf{Left:} The relationship between the mean of 3D AP40 across three classes
        with moderate difficulty (mAP) and the number of data calibration samples.
        \textbf{Right:} The relationship between the maximum observed input value in the
        first convolutional layer of PointPillars and the number of data calibration
        samples. Note that the x-axis is provided in log scale.
        }
    \label{fig:number-calib}
\end{figure}

For PTQ, we observe that using a very small number of calibration samples, $|D_{cal}|
\ll  |D_{train}|$ (e.g. four input frames), reduces the chance of encountering extreme
outliers, as demonstrated on the right side of Fig. \ref{fig:number-calib}. This
demonstrates a positive correlation between the maximum observed input value and the
number of calibration samples.

On the left side of Fig. \ref{fig:number-calib}, INT8 PointPillars with the first layer
kept as FP32 is used to demonstrate a negative correlation between the mean 3D AP40
across three classes with moderate difficulty (mAP) and the number of data calibration
samples. This indicates that PointPillars is more sensitive to rounding error than
clipping error. The difference in mAP between four input frames and 4,096 frames is
23.82\%. Therefore, throughout this work, only four input frames are used for the PTQ
data calibration.

\section{Experimental Results and Discussion}
\label{sec:experimental}
In this section, to demonstrate our proposed method, we conducted experiments with KITTI
3D object detection dataset \cite{geiger2012we}. \texttt{MMDetection3D}
\cite{mmdet3d2020} was used to train and evaluate our PointPillars models. The default
training setting of \texttt{MMDetection3D} for PointPillars was used, with the exception
that the learning rate for QAT was adjusted to $2 \times 10^{-4}$. Additionally, to
ensure similar behaviors between \texttt{MMDetection3D} and TensorRT, \texttt{legacy}
option for \texttt{voxel\_encoder} was set to \texttt{False}.

To comply with TensorRT quantization schemes, we used \texttt{modelopt} \cite{modelopt}
version 0.39.0 to perform PTQ and QAT. \texttt{modelopt} was slightly modified to
support mixed precision models. FP32 pre-trained weights were used for the weight
initialization for both PTQ and QAT. To convert the PointPillars model to TensorRT, we
first converted our PyTorch \cite{paszke2019pytorch} models to ONNX \cite{bai2019} and
then converted the ONNX files to a TensorRT engine files using \texttt{trtexec}. We used
two TensorRT version 10.13.2.6 for RTX 4070Ti and 10.3.0 for Jetson Orin.

\subsection{Sensitive Layer Search}
To find the sensitivity of each layer, the pre-trained FP32 model was used as the base
model. Layer by layer, the layer within the base FP32 model was replaced with the
corresponding INT8 layer. Then, this model was calibrated with PTQ using four randomly
selected LIDAR frames from the training dataset and evaluated on the validation dataset.

\begin{figure}[htbp]
    \centerline{
        \includegraphics[scale=0.5]{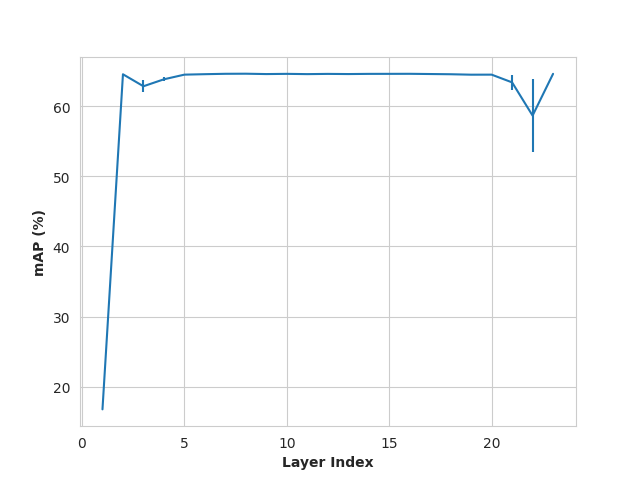}
    }
    \caption{
        Sensitive layer searches conducted across five random seeds, where errors bars
        indicate standard deviation. Y-axis indicates mAP. x-axis displays the layer
        index to assign as INT8 layer, where all other layers are FP32.
    }
    \label{fig:sensitive}
\end{figure}

To ensure that the sensitive layer search operates robustly across objects, the mAP was
used as the main metric. The results are visualized as shown in Fig. \ref{fig:sensitive}

From the figure, the top-3 most sensitive layers with the lowest mAP after INT8
conversion are voxel\_encoder.pfn\_layers.0.linear, bbox\_head.conv\_reg, and
backbone.blocks.0.3. The indices of these layers are 1, 22, and 3, as shown in Table
\ref{tab:latency}. These layers were chosen as the target layers for the mixed precision
model.

\subsection{Mixed Precision with PTQ and QAT}
After the sensitive layers were identified with PTQ, the top-k most sensitive layers
were selected to construct layer combinations for mixed precision models. We found that
only $k = 3$ is sufficient to provide the model performance close to the FP32 model.
Using a greedy search, top-1 to top-3 layers were used to create mixed precision models.
Subsequently, these mixed precision models were either re-calibrated again with PTQ or
trained with QAT to minimize the performance degradation.

For the notation of our mixed precision models, most layers are quantized to INT8, hence
only a few FP16 layers require to be described. Therefore, our notation of "FP16: x,y"
is used to describe a model with $x$-th and $y$-th as FP16 layers and all other layers
are INT8 layers.

After tuning the mixed precision models with either PTQ or QAT, the performances of our
models after TensorRT conversion are shown in Table \ref{tab:mixed-ptq-qat}.

\begin{table*}[htbp]
    \caption{
        Performance of baseline and mixed precision models with either PTQ or QAT. All
        models were converted to TensorRT format and evaluated for AP40.
    }
    \begin{center}
    \begin{tabular}{|c|c|c|c|c|c|c|c|c|c|c|c|}
        \hline
        \textbf{Data Type} &
        \multicolumn{3}{c|}{\textbf{Car AP40}} &
        \multicolumn{3}{c|}{\textbf{Cyclist AP40}} &
        \multicolumn{3}{c|}{\textbf{Pedestrian AP40}} & \textbf{mAP} \\
        \cline{2-10}
        & \textbf{Easy} & \textbf{Moderate} & \textbf{Hard}
        & \textbf{Easy} & \textbf{Moderate} & \textbf{Hard}
        & \textbf{Easy} & \textbf{Moderate} & \textbf{Hard} &
        \\
        \hline
        FP32 & 87.11 & 78.49 & \textbf{75.84} & 83.81 & \textbf{64.89} & 60.92 & 56.85 &
        50.55 & 46.05 & \textbf{64.64} \\
        \hline
        FP16 & 87.02 & 78.49 & 75.82 & 83.57 & 64.71 & \textbf{61.03} & \textbf{56.98} &
        50.53 & \textbf{46.22} & 64.58 \\
        \hline
        PTQ INT8 & 54.26 & 37.00 & 32.39 & 1.21 & 0.64 & 0.62 & 2.22 & 1.71 & 1.75 &
        13.12 \\
        \hline
        QAT INT8 & 83.25 & 72.32 & 67.70 & 78.28 & 56.82 & 53.35 & 51.47 & 45.58 & 42.03
        & 58.24 \\
        \hline
        \hline

        PTQ FP16: 1 & 83.84 & 74.68 & 70.30 & 78.61 & 61.46 & 57.81 & 54.73 & 47.94 &
        43.04 & 61.36 \\
        \hline
        PTQ FP16: 1,22 & 87.29 & 75.34 & 70.48 & 81.20 & 60.88 & 57.09 & 56.07 & 50.42 &
        46.04 & 62.21 \\
        \hline
        PTQ FP16: 1,22,3 & 86.78 & 77.52 & 73.22 & 82.20 & 63.78 & 59.72 & 56.32 & 50.23
        & 45.94 & 63.84 \\
        \hline
        \hline

        QAT FP16: 1 & 86.32 & 75.51 & 72.90 & 77.87 & 59.74 & 55.85 & 55.16 & 48.97 &
        45.08 & 61.40 \\
        \hline
        QAT FP16: 1,22 & \textbf{87.55} & 76.64 & 73.90 & \textbf{83.92} & 63.79 & 59.78
        & 54.13 & 48.20 & 43.69 & 62.88 \\
        \hline
        QAT FP16: 1,22,3 & 87.49 & \textbf{78.58} & 75.80 & 81.98 & 64.01 & 60.47 &
        56.97 & \textbf{50.83} & 46.03 & 64.47 \\
        \hline
    \end{tabular}

    \label{tab:mixed-ptq-qat}
    \end{center}
\end{table*}

With mixed precision layers and a very small number of data calibration, our mixed
precision PTQ models enhance the performance of the baseline PTQ model, "PTQ INT8". By
assigning only one sensitive layer to FP16 with PTQ, "PTQ FP16: 1" outperforms "PTQ
INT8" with mAP of 48.24\%.

In Table \ref{tab:mixed-ptq-qat}, there is a trend where replacing more sensitive layers
with FP16 reduces performance degradation. Even without retraining, "PTQ FP16: 1,22,3"
approaches the FP32 model, achieving an mAP of 63.84\%, while "QAT FP16: 1,22,3 "
provides an mAP of 64.47\%

Furthermore, QAT provides an improvement over PTQ, ensuring a smaller performance gap
between QAT and FP32. For instance, when comparing "PTQ FP16: 1,22,3" and "QAT FP16:
1,22,3", QAT provides an mAP improvement of 0.63\% over PTQ and is only 0.17\% lower
than the FP32 model.

\subsection{Runtime Latency}
\label{sec:latency}
In this section, we used \texttt{trtexec} to benchmark the runtime of our TensorRT
models, as shown in Table \ref{tab:latency}.

\begin{table}[htbp]
    \caption{Mean latency per devices across datatypes}
    \begin{center}
    \begin{tabular}{|c|c|c|c|}
        \hline
        \textbf{Data}&\multicolumn{2}{|c|}{\textbf{Mean Latency (ms)}} \\
        \cline{2-3}
        \textbf{Type} & \textbf{Jetson Orin} & \textbf{RTX 4070Ti} \\
        \hline
        FP32 & 32.91 & 3.713 \\
        \hline
        FP16 & 18.27 & 1.866 \\
        \hline
        INT8 & 14.77 & 1.774 \\
        \hline
        \hline
        FP16: 1 & \textbf{14.29} & \textbf{1.463} \\
        \hline
        FP16: 1,22 & 17.91 & 1.797 \\
        \hline
        FP16: 1,22,3 & 18.40 & 1.837 \\
        \hline
    \end{tabular}
    \label{tab:latency}
    \end{center}
\end{table}

Our model provides lower latency than other models with "FP16: 1", and compared to the
FP32 model, "FP16: 1" reduces latency by a factor of 2.538 on the RTX4070Ti. However,
adding more FP16 layers results in higher latency. We attribute this to the additional
quantization and dequantization are required to convert between FP16 and INT8 datatypes.
However, with RTX 4070Ti, this issue becomes less significant, as "FP16: 1,22,3" with
three FP16 layers still provides lower latency than FP16 models.

\subsection{Correlations between Number of Calibration Samples and Easy and Hard Samples}
\label{sec:corr}
We propose using a very small number of calibration samples to prioritize minimizing
rounding error over the clipping error. While this improves overall performance, as
shown in Fig. \ref{fig:number-calib}, it raises the question of whether this very small
sample size will asymmetrically affect Hard samples (with smaller bounding boxes, higher
occlusion levels and truncation) compared to Easy and Moderate samples or not.

\begin{figure}[htbp]
    \centerline{
        \includegraphics[scale=0.5]{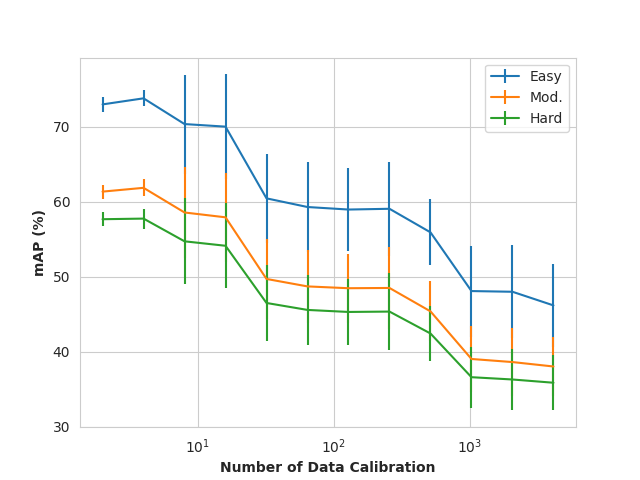}
    }
    \caption{
        The relationship between mAP and the number of data calibration samples across
        three difficulty levels.
    }
    \label{fig:diff}
\end{figure}

We investigated this question by measuring the correlations between the number of
calibration samples and mAP across three difficulty levels, as shown in Fig.
\ref{fig:diff}, using the same settings as Fig. \ref{fig:number-calib}. The results show
that the number of calibration samples exhibits a linear relationship with mAP across
all difficulties, with a Pearson correlation for each level greater than 0.998.
Therefore, the impact of reducing calibration samples is consistent across difficulty
levels, suggesting this method remains robust in high-risk scenarios.

\section{Conclusion}
\label{sec:conclusion}
In this work, we proposed a mixed precision framework by assigning FP to sensitive
layers, while other layers are assigned to INT8. Our framework discovers these sensitive
layers by assigning one layer at a time to INT8 with PTQ and evaluating average
precision. The top-$k$ layers with low average precision are assigned as the sensitive
layers. Then, the sensitive layers are greedy searched to formulate candidate models for
final PTQ or QAT.

With this framework, PTQ only pipeline promises competitive models without any training,
and QAT pipeline achieves the performance equivalent to FP32 models. We also
demonstrated that our models deliver lower latency compared to FP32
models by up to 2.538 times.

For future works, our works mainly support for FP32, FP16 and INT8 datatypes. We plan to
extend our framework to other datatype supported by modern GPUs: 4-bit integer (INT4),
8-bit floating point (FP8), and others. We also plan to extend this work to cover other
3D object detection models.

\section*{Acknowledgment}
\label{sec:acknowledgment}
We appreciate all anonymous reviewers for their reviews and comments to improve this
manuscript.

To enhance readability, all sections of this paper have been edited for
grammar using ChatGPT 5 \cite{chatgpt} and Gemini 3 \cite{gemini}.

This work was supported by Council for Science, Technology and Innovation (CSTI),
Cross-ministerial Strategic Innovation Promotion Program (SIP) Phase 3, Construction of
smart mobility platform, "Development of infrastructure and onboard sensor systems that
utilize compact LiDAR technology to understand the actual situations of streets in
living areas and busy districts" (funded by NEDO).
\bibliographystyle{IEEEtran}
\bibliography{reference}

\end{document}